\documentclass[a4paper,fleqn]{cas-sc}

\usepackage[authoryear]{natbib}
\usepackage{amsmath}
\usepackage{float}
\usepackage{graphicx}
\usepackage{atbegshi}
\usepackage{url}
\def\tsc#1{\csdef{#1}{\textsc{\lowercase{#1}}\xspace}}
\tsc{WGM}
\tsc{QE}
\tsc{EP}
\tsc{PMS}
\tsc{BEC}
\tsc{DE}

\begin{document}
\let\WriteBookmarks\relax
\def\floatpagepagefraction{1}
\def\textpagefraction{.001}

\title [mode = title]{A hybrid approach to Real-Time Multi-Target Tracking}


%
\author[1,2]{VM Scarrica}[type=editor,
                        auid=000,bioid=1,
                        prefix=Mr.,
                        role=PhD. Student,
                        orcid=0009-0008-4640-2693]

\cormark[1]


\ead{vincenzomariano.scarrica001@studenti.uniparthenope.it}

\author[2]{C Panariello}[                     
                        prefix=Mr.
                        ]

\ead{ciro.panariello001@studenti.uniparthenope.it}           

\author[2]{A Ferone}[%
   prefix=Prof.
   ]

\ead{alessio.ferone@uniparthenope.it}

\affiliation[1]{organization={National PhD Program in AI – Agrifood and Environment, University of Naples Federico II,  (Italy)},
    addressline={Corso Umberto I 40 }, 
    city={Naples},
    postcode={80138}, 
    country={Italy}} 

\author[2]{A Staiano}[%
   prefix=Prof.,
   role=Co-ordinator
   ]

\ead{antonino.staiano@uniparthenope.it}

\affiliation[2]{organization={University of Naples Parthenope, Department of Science and Technology},
    addressline={Centro Direzionale Isola C4}, 
    city={Naples},
    postcode={80143}, 
    country={Italy}}

\cortext[cor1]{Corresponding author}

\begin{abstract}
Multi-Object Tracking, also known as Multi-Target Tracking, is a significant area of computer vision that has many uses in a variety of settings. The development of deep learning, which has encouraged researchers to propose more and more work in this direction, has significantly impacted the scientific advancement around the study of tracking as well as many other domains related to computer vision. In fact, all of the solutions that are currently state-of-the-art in the literature and in the tracking industry, are built on top of deep learning methodologies that produce exceptionally good results. Deep learning  is enabled thanks to the ever more powerful technology researchers can use to handle the significant computational resources demanded by these models. However, when real-time is a main requirement, developing a tracking system without being constrained by expensive hardware support with enormous computational resources is necessary to widen tracking applications in real-world contexts. To this end, a compromise is to combine powerful deep strategies with more traditional approaches to favor considerably lower processing solutions at the cost of less accurate tracking results even though suitable for real-time domains. Indeed, the present work goes in that direction, proposing a hybrid strategy for real-time multi-target tracking that combines effectively a classical optical flow algorithm with a deep learning architecture, targeted to a human-crowd tracking system exhibiting a desirable trade-off between performance in tracking precision and computational costs. The developed architecture was experimented with different settings, and yielded a MOTA of 0.608 out of the compared state-of-the-art 0.549 results, and about half the running time when introducing the optical flow phase, achieving almost the same performance in terms of accuracy. 
\end{abstract}


\begin{highlights}
\item A novel hybrid algorithm for multi-object tracking
\item Optical flow and deep learning junction to improve efficiency
\item Encouraging outcomes on MOT2015, with a MOTA of 0.608
\end{highlights}


\maketitle

\section{Introduction}
%
%
%
%
As one of the most relevant topics in computer vision in the last few years, multi-object tracking remains perhaps the hardest challenge to tackle because of its complex intrinsic reasoning \cite{mot_review}. The spectrum of applications where this task can be found is ample, and it comprises activities such as human interaction, intelligent monitoring, auto-piloting, virtual reality, surgical navigation, crime prediction, missile navigation, missile reconnaissance, and many others. In human interaction, actions like the motion of arms, of the head, or of fingers can be tracked by a camera in order to manage programs \cite{human_computer_interaction}; in intelligent monitoring, systems learn how to monitor objects in order to compute statistics \cite{intelligence_monitoring}; in auto-piloting (for instance, self-driving cars), visual object tracking of other vehicles can be accomplished for avoiding dangers and accidents \cite{automatic_driving}; in surgical navigation, as a concern, objects like lancets and pliers can be tracked for improving their usage during surgery; in crime prediction, suspicious people can be monitored for avoiding anomalies and crimes in particular situations (for example, in a bank chamber) \cite{video_surveillance}; in missile navigation and reconnaissance, intelligent weapons can be tracked for correcting their trajectory \cite{aereospace}. Although the history of multiple-object tracking has seen successful works with convolutional neural networks \cite{Ferone} and, more recently, with transformers, which provide fairly good performance on large data sets, newer techniques still suffer from major problems, mainly related to the type of video being fed to the system \cite{challenge_tracking}. Some of those issues are described as follows: 
\begin{itemize}
    \item \textit{scale variation}, that is, when objects in the scene change their perspective and the system struggles to adjust their bounding box shapes (or masks);
    \item \textit{illumination conditions and/or variations}, affect pixel intensity and reflections on the camera objective, impacting the tracking quality;
    \item \textit{occlusions}, that is target objects are occluded by other tracked objects or by background elements;  
    \item \textit{out of view} happens when target objects get out of the scene, and then they might come back after some frames (besides, they can present a different shape or texture).
\end{itemize}

Other transformations that can affect tracking are \textit{deformations} and \textit{in-plane}/\textit{out-of-plane rotations}. Two further phenomena that make it difficult to re-detecting target objects or keep their motion path are \textit{motion blurring} and \textit{fast motion}. Ambiguities in results can also be caused by \textit{background clutter}, which is made up of objects that are confused with the real target objects, while low-resolution targets can produce a high mismatch rate. 
It is therefore a fact that embedded systems for real-time applications must lower their expectations for very high accuracy in order to achieve reasonable compromises in real-time tracking performance. The critical point here is how the search for a good trade-off affects tracking accuracy by reducing detection capabilities. 
The purpose of this work is to propose a solution that tries to limit the accuracy degradation while increasing the running performance of a multi-target real-time tracking system.  
The novelty lies in experiencing a state-of-the-art architecture to determine the most expensive components in terms of runtime and to replace these elements with methods that guarantee runtime gains at a low cost of accuracy degradation.
The rest of the paper is organized as follows: Section \ref{two} briefly overviews the tasks of a tracking system and reviews the state-of-the-art literature; Section \ref{three} deepens the building blocks of the architecture and describes how it was made hybrid with the combined use of the optical-flow technique; Section \ref{four} shows the performance of the proposed systems compared to the SOTA techniques on a combination of several benchmark datasets, and finally, some concluding remarks close the paper.

\section{Tracking tasks and Related works}\label{two}
The goal of a tracking system, given a sequence of semantically consecutive video frames, is to be able to identify target objects of interest, maintain their identity, and provide their individual trajectories across the considered frames \cite{mot_review}.
Object tracking can be categorized into two macro-tasks, namely, single-object tracking (SOT) and multiple-object tracking (MOT). In SOT, the tracking algorithm is applied to only one object in the scene, even though, within the scene, there can be multiple objects, even of the same class, whereas in MOT, more than one object (usually, from the same class, but not necessarily) in the scene is tracked at the same time. Moreover, in SOT, the focus is on the quality of predictions, while in MOT, since the algorithms have to deal with many instances, they pay more attention to the quantity and efficiency of parallel processing. This distinction is crucial since the two categories follow different approaches. In SOT, it is used to study traditional methods like correlation filters, optical flow methods \cite{optical_flow}, object detectors \cite{tracking_by_detection}, involving mostly Convolutional Neural Networks (CNN) \cite{cnn_intro}, although currently, they are moving to solutions involving \textit{Siamese network}-based \cite{siamese_network} and transformer-based methods \cite{transformers}. MOT algorithms have usually involved a well-known pipeline, starting from a detection algorithm \cite{tracking_by_detection2}, then an association algorithm for assigning each predicted bounding box (or mask) to a specific object or entity instance, and finally a motion estimation phase. \\
Both SOT and MOT algorithms can then be divided into \textit{short-term}, where the target object never disappears from the scene, and \textit{long-term}, i.e. when an object can leave a scene and reappear after a certain number of frames. Obviously, the latter case is more complex because the object re-entering the scene has to be properly re-identified. Finally, we can distinguish between \textit{online} and \textit{offline} algorithms. Online procedures do not take into account external information for training the tracking algorithm to improve its performance and are only based on the information from previous frames. Instead, offline procedures can exploit external data for training and inference. \\
In the literature, MOT pipelines have been studied for a long time, and some famous algorithms have been widely used for each step: in the object detection step, there are models like Faster-RCNN \cite{faster_rcnn}, YOLO \cite{yolo}, and Mask-RCNN \cite{maskrcnn}; in the data association step, there is the Hungarian algorithm \cite{algoritmo_ungherese} and the Byte methodology \cite{byte_track}. At the motion estimation step, mainly the Kalman filter \cite{kalman_filter} is employed. The SORT algorithm has combined some of the above techniques, it uses Faster R-CNN as an object detector, the Hungarian algorithm for data association using the intersection over union (IoU), and a Kalman filter for motion estimation \cite{sort}. An advanced version of the latter work is DeepSORT, which extracts features from each bounding box of a CNN for the re-identification phase (instead of using the IoU metric) and then uses cosine distance \cite{deep_sort}. Both SORT and DeepSORT are two-stage algorithms because they separate the re-identification from the data association using two feed-forward steps. Other approaches use fixed anchors for object segmentation applied to multi-object tracking, such as Track-RCNN \cite{trackrcnn}.\\ FairMOT is an online end-to-end multi-object tracker; it combines two steps into one, speeding up the tracking process \cite{fairmot_paper}. The big difference with previous state-of-the-art models is that it can be trained by providing one image at a time, not necessarily two images, thanks to its online nature. It is based on CenterNet \cite{centernet}, where the architecture is composed of an encoder-decoder model (consisting of ResNet-34 for feature extraction \cite{resnet_paper}) and two homogeneous branches for detection and re-identification. An advantage of this method is the execution time, which is lower than the previous works in terms of FPS; a major disadvantage is in the accuracy, which admits some leakage due to the use of fixed anchors. \\
MOT has also been tackled using optical flow methods, specifically, the Pyramidal Lucas Kanade feature tracker \cite{lk_piramide}. This algorithm generates a pyramid of the same image with different smaller resolutions, where objects get closer to each other. This process makes the search easier and faster.\\ 
Aiming to develop an efficient system for real-time domains, with a good trade-off between accuracy and computation time, this paper proposes an approach that combines two different methodologies: on the one hand, a deep learning-based architecture based on FairMOT with newly introduced modifications that will be discussed in the next section, and on the other hand, an optical flow-based framework based on Pyramidal Lucas Kanade. The hybrid approach was conceived to speed up the boundary boxes computation and the re-identification step of the original FairMOT model.

\section{Methodology}\label{three}
As mentioned above, the proposed architecture is based on the FairMOT model.
In the following, the basic characteristics of the model are reviewed, and then a reliable data association technique called Byte is introduced. Finally, the modifications to the original model leading to the proposed hybrid solution are presented.
\subsection{FairMOT}
\begin{figure}[!ht]
\centering
\includegraphics[width=.50\textwidth]{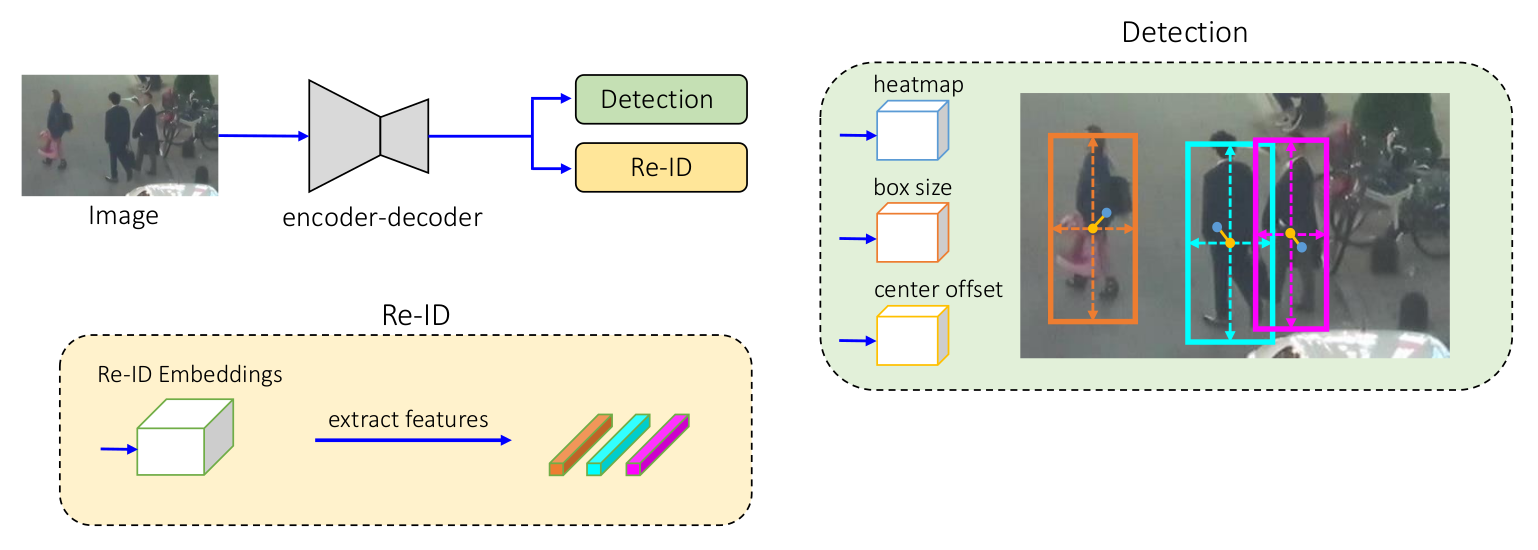}
\caption[FairMOT overview]{Overview of the FairMOT one-shot tracker. The input image is first sent over an encoder/decoder network to extract high-resolution features. After that, two homogeneous branches are added for object detection and feature re-id extraction. The feature vector in the center of the bounding box of the predicted object is used for the data association phase. The illustration is taken from \cite{fairmot_paper}.}
\label{fig:fairmot_overview}
\end{figure}

FairMOT (see fig. \ref{fig:fairmot_overview}) is based on CenterNet \cite{centernet}, an architecture composed of an encoder-decoder model consisting of a ResNet-34 for feature extraction \cite{resnet_paper}, which takes a $1088 \times 608$ input frame providing, as a result of the decoding phase, a feature map whose size is $1/4$ of the original image. The feature map goes into two branches, namely \textit{detection} and \textit{re-identification}.
The Kalman filter is used for motion estimation, where the Mahalanobis distance is used between the predicted bounding box positions and Kalman-estimated bounding box positions, while the cosine distance is used between the re-identified bounding box features and Kalman-estimated bounding box features. All distances are then combined and provided as a matrix to feed into the Hungarian data association algorithm.
The \textit{detection branch} outputs three heads—a heatmap, a center offset, and a bounding box—by taking features from the backbone as input. The center of each bounding box is located by the heatmap head, whose response is an exponential function that admits a maximum in the center of the box equal to $1$. The associated loss function for this head is described in eqn. \ref{eqn:losshead} and is defined as a \textit{pixel-wise logistic regression} with focal loss \cite{focal_loss}, where $M_{xy}$ is the initial response, $\hat{M}$ is the estimated heatmap, and $\alpha,\beta$ are the given parameters in the focal loss. 

\begin{equation}
L_{\text{heat}} = -\frac{1}{N} \sum_{xy} 
    \begin{cases}
        \left(1-\hat{M}_{xy}\right)^\alpha log \hat{M}_{xy}, \text{if } M_{xy}=1 \text{;}\\
        \left(1-\hat{M}_{xy}\right)^\beta \hat{M}^{\alpha}_{xy} log \left( 1 - \hat{M}_{xy} \right) \text{else}  
    \end{cases}
    \label{eqn:losshead}
\end{equation}

By providing an integer approximation of the center of the bounding box coordinates when divided by 4 in the output response, the center offset head improves center localization. 
The expected box dimensions, W and H, are determined by the bounding box head. 
The bounding box head loss is described in eqn. \ref{eqn:boundinghead}, where $\lambda_s$ is a regularizing coefficient for the second term, and it is set to $0.1$ as in the original work \cite{centernet}: 

\begin{equation}
    L_{\text{box}} = \sum^{N}_{i=1} \left \Vert o^i - \hat{o}^i \right \Vert_1 + \lambda_s \left \Vert s^i - \hat{s}^i \right \Vert_1.
    \label{eqn:boundinghead}
\end{equation}
\\
The \textit{re-identification branch} is responsible for finding the target in the new frame; it receives an output response from a classification task.
After recovering the ground truth box features, they are transferred to a dense layer and then mapped to a one-hot encoded vector using the softmax function (where each bit represents a class of the task).
The re-identification head loss is described in eqn. \ref{eqn:reidhead}, where $K$ is the total number of identities in the training data:
\begin{equation}
    L_{\text{identity}} = -\sum^{N}_{i=1}\sum^{k}_{K=1} L^i\left(k\right) \text{log}\left(p\left(k\right)\right).
    \label{eqn:reidhead}
\end{equation}
\\
The three detection head losses and the re-identification loss are summed to form the total loss.
The total loss is shown in eqn. \ref{eqn:total}, where $w_1$ and $w_2$ are trainable parameters.
\begin{equation}
    \begin{cases}
        L_{\text{detection}} = L_{\text{heat}} + L_{\text{box}}\\
    
        L_{\text{total}} = \frac{1}{2} \left( \frac{1}{e^{w_1}} L_{\text{detection}} + \frac{1}{e^{w_2}} L_{\text{identity}} + w_1 + w_2 \right)
    \end{cases}
     \label{eqn:total}
     \end{equation}
\subsection{Data association: Byte method}
Classical tracking-by-detection methods retain only detected bounding boxes that have confidence greater than the threshold value. This is because the right trade-off must exist. Accepting detection boxes with low confidence values will improve the detection rate (true positives) but will also introduce false positives. Understanding whether bounding boxes with very low confidence values should be removed or not is a very important issue. Indeed, even with a low confidence value, the object may still exist in the scene, and ignoring it would reduce the efficiency of the detection model and consequently also in tracking. 

\begin{figure}[ht]
\centering
\includegraphics[width=.50\textwidth]{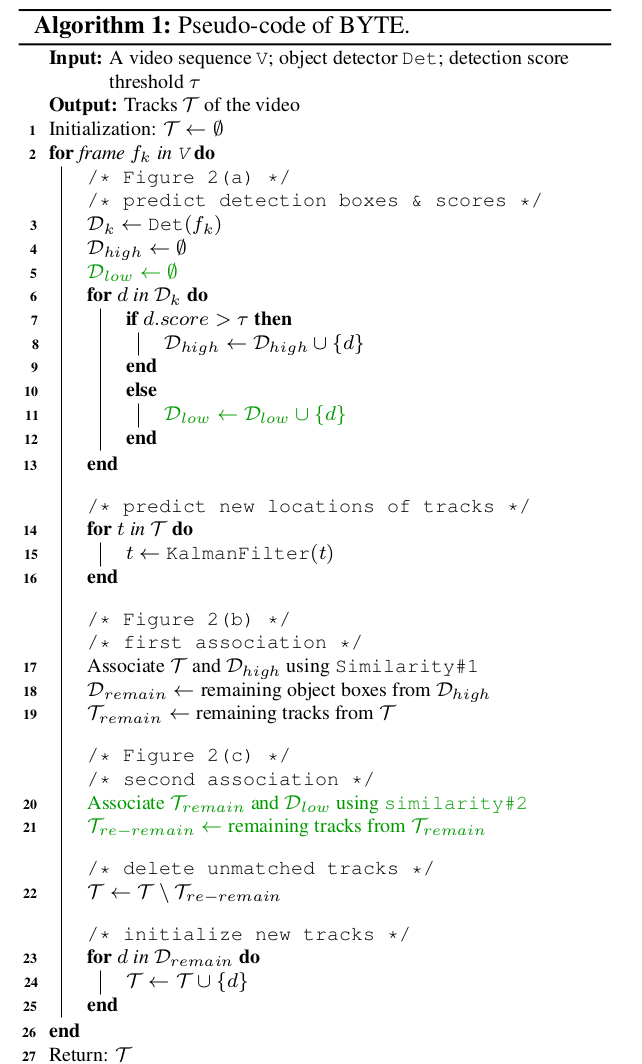}
\caption[Byte algorithm]{Byte pseudo-code. Relevant parts are evidenced in green. The illustration is taken from \cite{byte_track}.}
\label{fig:byte_algoritmo}
\end{figure}

Byte is a generic method for data association \cite{byte_track} that tries to solve the above difficulty. Using Byte (see the pseudo-code in fig. \ref{fig:byte_algoritmo}), all detected bounding boxes (both those above and below the threshold value) are maintained and handled, with likewise an efficient method for dealing with detection boxes with a low confidence value. 
As input, the algorithm takes a sequence of frames $V$, an object detector $Det$, and a threshold value for the detection phase $\tau$. It returns as output the list $T$ of objects tracked in each video frame. For each frame $f_k$ in $V$, the $Det$ algorithm is used to identify all object bounding boxes, which will be put into a $D_k$ list. Then, two lists are initialized, $D_{high}$ and $D_{low}$. For each bounding box $d$ in $D_k$, a check is made on the threshold value $\tau$. Concretely, if the confidence of $d$ has a value greater than $\tau$ this will be placed in $D_{high}$ otherwise in $D_{low}$. Thus $D_{high}$ contains all bounding boxes that have a confidence value greater than the threshold $\tau$ while $D_{low}$ contains all bounding boxes that have a lower confidence value. For each object in $T$ the Kalman filter is used to predict its new position. Now, there is a first stage of association between the currently tracked objects $T$ and the bounding boxes $D_{high}$ using the Hungarian algorithm. All bounding boxes that did not match in the association phase (e.g., new object entering the scene) are inserted into $D_{remain}$, and bounding boxes of the tracked objects that did not match (e.g., object in the scene disappearing) are inserted into $T_{remain}$. After that, a second phase of association is carried out between the objects in $T_{remain}$ and the objects with low detection confidence, which are present in $D_{low}$. In $T{re-remain}$ are stored the objects that did not find any matches also in this second phase of association. 
This is the heart of the method: Objects that did not match in the first association (i.e. have low confidence) have a chance to match since they are present in $D_{low}$. 
\subsection{Proposed modifications and hybrid approch}
 Several versions of the FairMOT architecture have been tested. Replacements of the encoding backbone were considered (ResNet-34 in the original paper). Two state-of-the-art models, namely ConvNeXt Tiny \cite{convnext} and EfficientNetB3 \cite{efficientnet}, were chosen to replace the original network. Then, in the decoding part, two alternative solutions were explored. The first proposal includes the use of deformable convolutions instead of transposed convolutions, so as to improve the generalization to the geometric transformation of the feature maps in the upsampling phase. The second proposal instead considers the use of Feature Pyramid Networks (FPN) \cite{fpn}, which adds transversal connections between the downsampling and upsampling stages, preserving more information about the shape and texture of the objects at a high level that is usually lost between the two stages. The data association method is replaced by the byte method, which tries to solve the problem of occluded target objects.

To speed up the entire tracking pipeline, a hybrid approach is presented that combines the single-image target initialization and re-identification procedures with an optical flow estimation approach using the Pyramidal Lucas Kanade (PLK) algorithm. The pipeline is as follows: after one step of a modified FairMOT, $N$ steps of PLK are applied, and then the process loops. The problem of the optical flow approach in terms of re-identification along the $N$ steps is twofold: key points of target objects are difficult to estimate, and bounding box estimation in the next frame is difficult to perform because key points change their positions. For the first problem, corner detection was applied (using the Features from Accelerated Segment Test (FAST) algorithm \cite{fast}), and for the second problem, the Random Samples Consensus (RANSAC) algorithm was used to estimate the transformation matrix of the key-points \cite{ransac}:

\begin{equation}
    \text{M} = 
    \begin{bmatrix}
        \text{cos}\left(\theta \right) \cdot s & \text{-sin}\left(\theta\right) \cdot s & t_x \\
        \text{sin}\left(\theta\right) \cdot s & \text{cos}\left(\theta\right) \cdot s & t_y\\
    \end{bmatrix}
    \label{eqn:matrix}
\end{equation}

In RANSAC, an affine application matrix $M$ is estimated (see eqn. \ref{eqn:matrix}), assuming as input the key-points of the target from the previous frame and the key-points of the target from the current frame:

\begin{equation}
    \begin{bmatrix}
        \text{x}_{new} \\ \text{y}_{new}
    \end{bmatrix}
    = \text{M} \cdot 
    \begin{bmatrix}
    x \\ y \\ 1 
    \end{bmatrix}
    \label{eqn:ransac}
\end{equation}

\begin{figure}[ht]
\centering
\includegraphics[width=.50\textwidth]{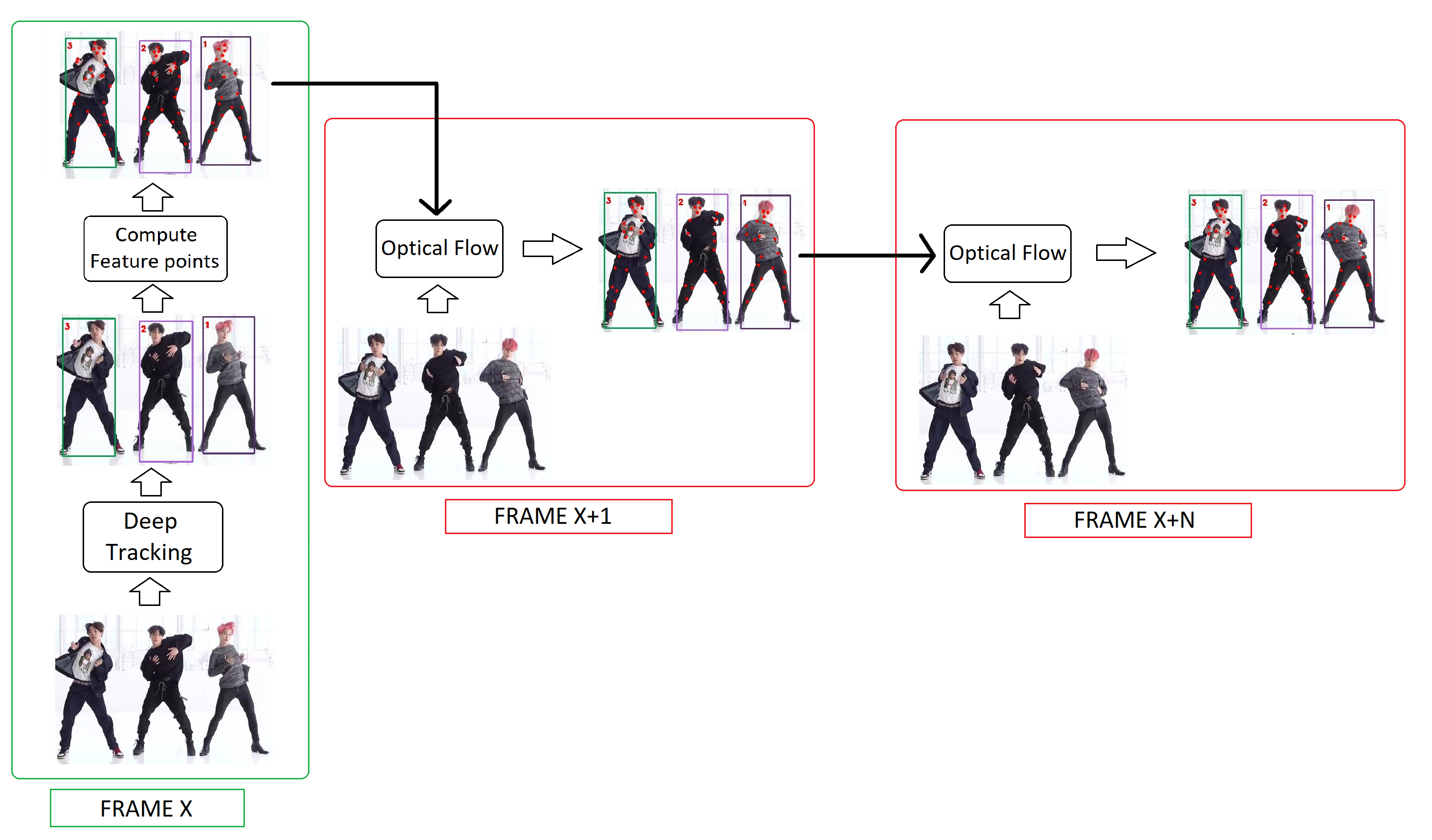}
\caption[Workflow]{A high-level view of the entire workflow. An input frame $X$ is processed by the FairMOT module in the Deep Tracking step. Then, feature points are computed using the FAST algorithm. These feature points are propagated to the following $N$ frames.
The subsequent $N$ frames are then submitted to the optical flow step, using the Pyramidal Lucas Kanade implementation. The optical flow at the $X+i$ frame takes as input the feature points of the previous frame and the current $X+i$-th frame, where $i < N$.
The samples are taken from the DanceTrack dataset \cite{sun2022dance}.
}
\label{fig:workflow}
\end{figure}

Then, by multiplying the first point on the left and the last point on the right of each bounding box by the corresponding matrix (as shown in eqn. \ref{eqn:ransac}), the new coordinates can be found easily and quickly. A problem of the whole pipeline is the setting of the hyperparameter $N$, which has been empirically optimized in this work. The whole pipeline is shown in fig. \ref{fig:workflow}.

The official GitHub repository of this work is freely available at \url{https://github.com/Dantekk/A-hybrid-approach-to-Real-Time-Multi-Target-Tracking}.
\section{Experiments}\label{four}
In this section, the experimental results are discussed, starting from the dataset to the parameter settings and the evaluation metrics.
\subsection{Dataset}
The experiments focused on tracking people as the main and only class of instances. The dataset used in the experiments is a mixture of several datasets representing scenes of human crowds. In details:
\begin{itemize}
\item CrowdHuman \cite{crowd_human_dataset} is a benchmark dataset useful for training and evaluating object detection algorithms on crowd scenarios.
It is large, rich in annotations, and extremely heterogeneous. It contains a total of 470,000 human instances, divided into training and validation sets, and an average density of 23 people per image, with different types of occlusions.
An excellent dataset, especially useful for training the recognition phase related to the detection of persons. 
\item CityPersons \cite{city_persons_dataset} is a subset of the CityScapes dataset, containing only images where people are present in street contexts, with an average density of 7 people per image.
\item CUHK-SYSU \cite{cuhs} is suitable for training the re-identification phase, as it contains images of people in different contexts and positions. It contains 18,184 pictures and 8,432 identities.
\item PRW \cite{prw_dataset} is an excellent dataset for training both the recognition phase and the person re-identification phase, as it contains 11,836 images and 932 identities of people in various scenarios, taken from six cameras with different shots.
\item ETHZ \cite{eth_biwi_00534} is a dataset suitable for tracking in road contexts. It contains about 2,804 images in 3 video sequences. The images were captured by a camera mounted on a car.
\item MOT2015 \cite{mot15} is a benchmark dataset for multiple object tracking, consisting of 11 different indoor and outdoor scenes of public places with pedestrians, with high variance in camera motion, camera angle, and imaging conditions. The average density of people for each frame is 7. It contains 5783 images.
\end{itemize}
All data sets except for MOT2015 have been employed for training phase. The total number of images is $61490$. The data is divided into $50422$ training images and $11068$ validation images. MOT2015 has been used for testing phase. The resolution of all images is the same as the input size of the models.

\subsection{Parameters setting and metrics}
'Parameters were organized as follows: the batch size was set to 12 for models using ConvNeXt Tiny and 16 for EfficientNetB3, respectively; the number of epochs was set to 50 and Adam was used as the optimizer; the learning rate was set to $10^{-4}$ and three learning rate schedulers were tried: fixed, step, and exponential. The best scheme was found with the step scheduler, as in the original paper. Online data augmentation was used during training, involving transformations such as rotations, scaling and jittering.
The metrics considered in the experiments are Multiple Object Tracking Accuracy (MOTA) and Multiple Object Tracking Precision (MOTP) \cite{mot15}.
MOTA can be defined as

\begin{equation}
 MOTA = 1 - \frac{\sum_{t}^{}FP_{t}+FN_{t}+IDS_{t}}{\sum_{t}^{}GT_{t}},
\end{equation}

where FP denotes false positives and FN false negatives, IDS is the mismatch error and GT is the target bounding box that should be in the frame.
It is between minus infinity and 1 and the best accuracy is achieved when it is equal to 1.
MOTP can be defined as

\begin{equation}
    MOTP = \frac{\sum_{i,t}D_{i,t}}{\sum_{t}C_{t}},
\end{equation}

where D is the distance between the predicted and target bounding boxes, and C is the total number of matches between the predicted and target boxes.
It ranges from 0 to 1. The best precision is achieved when it is equal to 0.

The system was implemented on a GPU (NVLINK) NVIDIA Tesla V100, 32GB SXM2 on clusters provided by the Department of Science and Technology at the University of Naples "Parthenope" for the training phase, while GPU NVIDIA GeForce RTX 3060 Ti has been used for inference on MOT15.

\subsection{Results}
To compare the performance between a standard MOT pipeline and the proposed hybrid solution, two sets of experiments were performed. Table \ref{tbl:tempi-di-esecuzione} shows the final results, including average running times, for a regular pipeline, while Table \ref{tbl:tempi-di-esecuzione-hybrid} shows the results for the hybrid approach.

\begin{table}[h!]
\centering
\begin{tabular}{|p{1.5cm}| c| c| c| c | c |}
        \hline
        \textbf{Model}  & \textbf{Byte} & \textbf{MOTA} & \textbf{MOTP} & \textbf{\# Parameters} & \textbf{FPS}\\
        \hline
        \hline
        ResNet-34 & no & 0.535 & 0.21 & 31.5M & $\sim$15\\
        \hline
        \textbf{ResNet-34} & \textbf{yes} & \textbf{0.549} & 0.209 &  31.5M & $\sim$\textbf{15}\\
        \hline
        ConvNeXt Tiny & no & 0.578 & 0.231 & 32M & $\sim$15\\
        \hline
        ConvNeXt Tiny & yes & 0.587 & 0.225 & 32M & $\sim$15\\
        \hline
        ConvNeXt Tiny FPN & no & 0.593 & 0.23 & 39M & $\sim$14/15\\
        \hline
        \textbf{ConvNeXt Tiny FPN} & \textbf{yes} & \textbf{0.608} & 0.224 & 39M & $\sim$\textbf{14/15}\\
        \hline
        EfficientNet B3 & no & 0.559 & 0.248 & 15.5M & $\sim$19\\
        \hline
        \textbf{EfficientNet B3} & \textbf{yes} & \textbf{0.567} & 0.243 & 15.5M & $\sim$\textbf{19}\\ 
        \hline
\end{tabular}
\caption[Running time measured in frame per second (FPS)]{Metrics and average running time in FPS of the regular approach on MOT15 using \textit{NVIDIA GeForce RTX 3060 Ti}.}
\label{tbl:tempi-di-esecuzione}
\end{table}

\begin{table}[h!]
\centering
\begin{tabular}{|p{1.5cm}| c| c| c| c | c |}
        \hline
        \textbf{Model} & \textbf{Skip} & \textbf{MOTA} & \textbf{MOTP} & \textbf{\# Parameters} & \textbf{FPS}\\
        \hline
        \hline
        \textbf{ResNet-34}  & \textbf{0} &\textbf{0.549} & 0.209 &  31.5M & $\sim$\textbf{15}\\
        \hline
        ConvNeXt Tiny FPN & 0 & 0.608 & 0.224 & 39M & $\sim$14/15\\
        \hline
        ConvNeXt Tiny FPN & 1 & 0.586 & 0.228 & 39M & $\sim$20\\
        \hline
        \textbf{ConvNeXt Tiny FPN} & \textbf{2} & \textbf{0.546} & 0.231 & 39M & $\sim$\textbf{28}\\
        \hline
        \textbf{ConvNeXt Tiny FPN} & \textbf{3} & \textbf{0.504} & 0.234 & 39M & $\sim$\textbf{35}\\
        \hline
        ConvNeXt Tiny FPN & 4 & 0.441 & 0.241 & 39M & $\sim$42\\
        \hline
        \hline
        EfficientNet B3  & 0 & 0.567 & 0.243 & 15.5M & $\sim$19\\ 
        \hline
        \textbf{EfficientNet B3}  & \textbf{1} & \textbf{0.539} & 0.245 & 15.5M & $\sim$\textbf{24}\\ 
        \hline
        \textbf{EfficientNet B3}  & \textbf{2} & \textbf{0.486} & 0.249 & 15.5M & $\sim$\textbf{32}\\ 
        \hline
        EfficientNet B3  & 3 & 0.447 & 0.254 & 15.5M & $\sim$39\\ 
        \hline
        EfficientNet B3  & 4 & 0.390 & 0.258 & 15.5M & $\sim$47\\ 
        \hline
\end{tabular}
\caption[Running time of hybrid approach]{Metrics and average running time in FPS of the hybrid approach on MOT15 using \textit{NVIDIA GeForce RTX 3060 Ti}. Skip stands for the number of frames subjected to optical flow estimation.}
\label{tbl:tempi-di-esecuzione-hybrid}
\end{table}

Table \ref{tbl:tempi-di-esecuzione} considers variants of the FairMOT model. The best MOTA result, $0.608$,  is obtained using ConvNeXt Tiny as the backbone and FPN as the decoder with Byte data association. The model using EfficientNetB3 gives a MOTA of 0.567, which is lower than the ConvNeXt-based model, but higher than the original paper ($0.549$) using ResNet-34. In terms of running time, the average FPS is around 15, except for the lighter models with EfficientNetB3, where the average FPS increases to $19$. In all cases, the Byte data association method improves MOTA and MOTP (when the backbone is ResNet-34). MOTP shows no improvement in the other experiments and is set to deteriorate.

\begin{figure}[ht]
\centering
\includegraphics[width=0.21\textwidth]{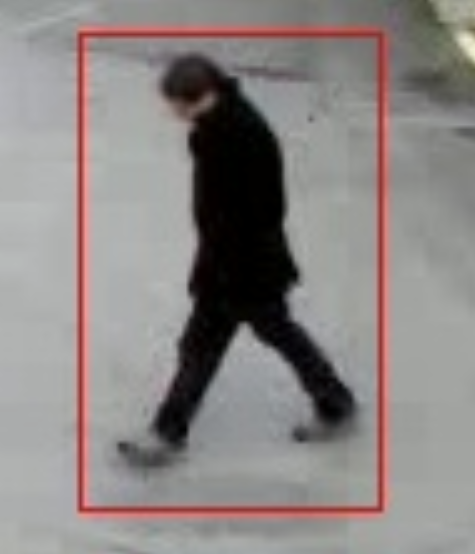}{(a)}
\includegraphics[width=0.21\textwidth]{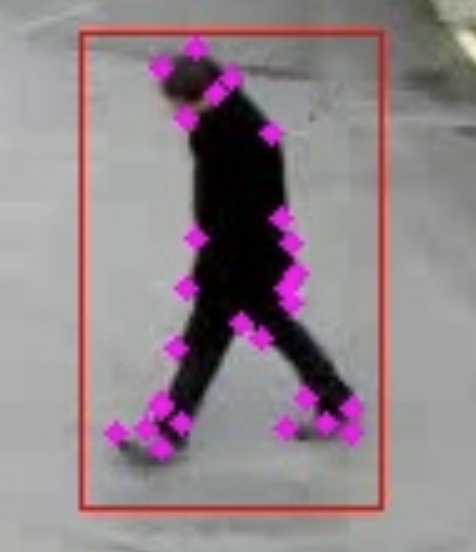}{(b)}
\includegraphics[width=0.21\textwidth]{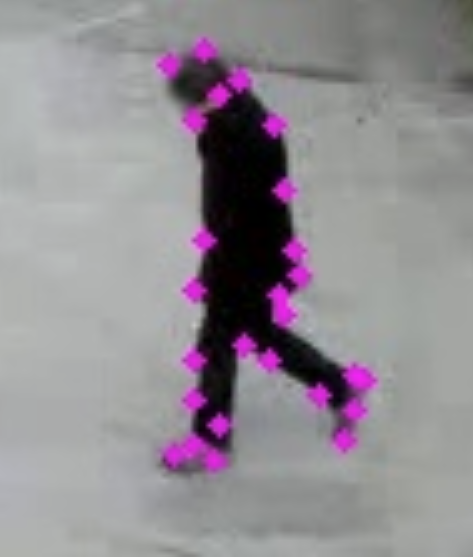}{(c)}
\includegraphics[width=0.21\textwidth]{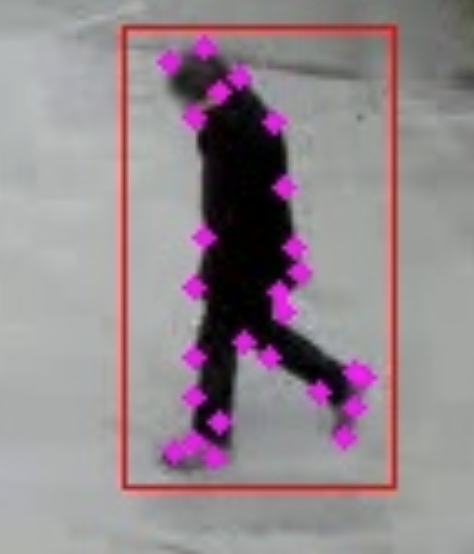}{(d)}
\caption{(a) Bounding box detected in the detection step. (b) Features extracted from the object. (c) New key-points estimated through the PLK method. (d) Bounding box obtained from the key points using the RANSAC method. }
\label{img:immagine}
\end{figure}

According to Table \ref{tbl:tempi-di-esecuzione-hybrid}, instead, no improvements in terms of MOTA or MOTP have been introduced by the hybrid approach, but results boost on running time. In the best experiment, using ConvNeXt Tiny with FPN and a number of optical flow skips set to 2, the MOTA is approximately the same as in the original paper (using ResNet-34), but with a higher FPS value, 28 instead of 15. When using EfficientNetB3, the best MOTA results are found with a number of skips set to 1 or 2 (0.539 and 0.486, respectively), but these results do not beat the original ones. MOTP does not get any improvements with the hybrid approach compared to the original paper.
In Fig. \ref{img:immagine}, the processing of two frames is shown: the first two ((a) and (b)) show the first frame detection with FairMOT ConvNeXt Tiny FPN, while the third and fourth ((c) and (d)) show the second frame where optical flow has been executed and key-points are computed. In (d), the bounding box is estimated using RANSAC.

\label{s2}
\section{Conclusions}
The goal of this work was to develop a system for tracking multiple people in real-time using hardware support with limited computing resources. To achieve this goal, it was necessary to develop an architecture that would find the right compromise between performance and required computational resources. For this purpose, the deep FairMOT model was chosen as the base architecture. This model belongs to the tracking-by-detection typology and represents a turning point for MOT models. In fact, it is a \textit{one-stage tracker}, i.e. in a single feed-forward step, it can detect the objects present in the scene and compute the re-ID features necessary for the association phase. FairMOT has been improved in several ways. First, the Byte Methodology was introduced into the FairMOT model. This improved the \textit{data association} phase and the model results in general. Next, the encoder/decoder structure underlying FairMOT was modified. On the encoder side, the ResNet-34 architecture was replaced by ConvNeXt Tiny and EfficientNet B3. On the decoder side, improvements were made using FPNs, lateral connections, and deformable convolutions. Overall, these modifications improved the already good performance of the original model. Finally, to guarantee real-time performance, a hybrid approach combining deep learning and the Lucas-Kanade method for optical flow was developed. In this way, it was possible to find the right trade-off between performance and computational speed, thus ensuring good real-time tracking performance with relatively limited hardware. The proposed system has been tested on the MOT15 dataset, and the results are quite encouraging and could be used in several areas, from video surveillance to crime prediction and workplace monitoring. \\
Future developments could consider other variants of the coding phase, for example using transformers as well as ViT or Swin, or changing the object detector from CenterNet to other popular algorithms as well as the YOLO family. Finally, hard re-identification features could be studied, covering the cases where objects re-enter a scene with different transformations, shapes, or textures.

\section{CRediT authorship contribution statement}
All authors contributed equally to this work.
\printcredits
\bibliographystyle{cas-model2-names}

\bibliography{cas-refs}

\begin{thebibliography}{40}
\expandafter\ifx\csname natexlab\endcsname\relax\def\natexlab#1{#1}\fi
\providecommand{\url}[1]{\texttt{#1}}
\providecommand{\href}[2]{#2}
\providecommand{\path}[1]{#1}
\providecommand{\DOIprefix}{doi:}
\providecommand{\ArXivprefix}{arXiv:}
\providecommand{\URLprefix}{URL: }
\providecommand{\Pubmedprefix}{pmid:}
\providecommand{\doi}[1]{\href{http://dx.doi.org/#1}{\path{#1}}}
\providecommand{\Pubmed}[1]{\href{pmid:#1}{\path{#1}}}
\providecommand{\bibinfo}[2]{#2}
\ifx\xfnm\relax \def\xfnm[#1]{\unskip,\space#1}\fi
\bibitem[{Andriluka et~al.(2008)Andriluka, Roth and
  Schiele}]{tracking_by_detection}
\bibinfo{author}{Andriluka, M.}, \bibinfo{author}{Roth, S.},
  \bibinfo{author}{Schiele, B.}, \bibinfo{year}{2008}.
\newblock \bibinfo{title}{People-tracking-by-detection and
  people-detection-by-tracking}, in: \bibinfo{booktitle}{2008 IEEE Conference
  on Computer Vision and Pattern Recognition}, pp. \bibinfo{pages}{1--8}.
\newblock \DOIprefix\doi{10.1109/CVPR.2008.4587583}.
\bibitem[{Bewley et~al.(2016)Bewley, Ge, Ott, Ramos and Upcroft}]{sort}
\bibinfo{author}{Bewley, A.}, \bibinfo{author}{Ge, Z.}, \bibinfo{author}{Ott,
  L.}, \bibinfo{author}{Ramos, F.T.}, \bibinfo{author}{Upcroft, B.},
  \bibinfo{year}{2016}.
\newblock \bibinfo{title}{Simple online and realtime tracking}.
\newblock \bibinfo{journal}{2016 IEEE International Conference on Image
  Processing (ICIP)} , \bibinfo{pages}{3464--3468}.
\bibitem[{Bouguet(2000)}]{lk_piramide}
\bibinfo{author}{Bouguet, J.Y.}, \bibinfo{year}{2000}.
\newblock \bibinfo{title}{Pyramidal implementation of the lucas kanade feature
  tracker description of the algorithm}.
\newblock \URLprefix
  \url{http://robots.stanford.edu/cs223b04/algo_tracking.pdf}.
\bibitem[{Duan et~al.(2019)Duan, Bai, Xie, Qi, Huang and Tian}]{centernet}
\bibinfo{author}{Duan, K.}, \bibinfo{author}{Bai, S.}, \bibinfo{author}{Xie,
  L.}, \bibinfo{author}{Qi, H.}, \bibinfo{author}{Huang, Q.},
  \bibinfo{author}{Tian, Q.}, \bibinfo{year}{2019}.
\newblock \bibinfo{title}{Centernet: Keypoint triplets for object detection},
  in: \bibinfo{booktitle}{2019 IEEE/CVF International Conference on Computer
  Vision (ICCV)}, pp. \bibinfo{pages}{6568--6577}.
\newblock \DOIprefix\doi{10.1109/ICCV.2019.00667}.
\bibitem[{Ess et~al.(2008)Ess, Leibe, Schindler and Van~Gool}]{eth_biwi_00534}
\bibinfo{author}{Ess, A.}, \bibinfo{author}{Leibe, B.},
  \bibinfo{author}{Schindler, K.}, \bibinfo{author}{Van~Gool, L.},
  \bibinfo{year}{2008}.
\newblock \bibinfo{title}{A mobile vision system for robust multi-person
  tracking}, in: \bibinfo{booktitle}{2008 IEEE Conference on Computer Vision
  and Pattern Recognition}, pp. \bibinfo{pages}{1--8}.
\newblock \DOIprefix\doi{10.1109/CVPR.2008.4587581}.
\bibitem[{Fischler and Bolles(1981)}]{ransac}
\bibinfo{author}{Fischler, M.A.}, \bibinfo{author}{Bolles, R.C.},
  \bibinfo{year}{1981}.
\newblock \bibinfo{title}{Random sample consensus: A paradigm for model fitting
  with applications to image analysis and automated cartography}.
\newblock \bibinfo{journal}{Commun. ACM} \bibinfo{volume}{24},
  \bibinfo{pages}{381–395}.
\newblock \URLprefix \url{https://doi.org/10.1145/358669.358692},
  \DOIprefix\doi{10.1145/358669.358692}.
\bibitem[{He et~al.(2017)He, Gkioxari, Doll{\'{a}}r and Girshick}]{maskrcnn}
\bibinfo{author}{He, K.}, \bibinfo{author}{Gkioxari, G.},
  \bibinfo{author}{Doll{\'{a}}r, P.}, \bibinfo{author}{Girshick, R.B.},
  \bibinfo{year}{2017}.
\newblock \bibinfo{title}{Mask {R-CNN}}.
\newblock \bibinfo{journal}{CoRR} \bibinfo{volume}{abs/1703.06870}.
\newblock \URLprefix \url{http://arxiv.org/abs/1703.06870},
  \href{http://arxiv.org/abs/1703.06870}{\tt arXiv:1703.06870}.
\bibitem[{He et~al.(2016)He, Zhang, Ren and Sun}]{resnet_paper}
\bibinfo{author}{He, K.}, \bibinfo{author}{Zhang, X.}, \bibinfo{author}{Ren,
  S.}, \bibinfo{author}{Sun, J.}, \bibinfo{year}{2016}.
\newblock \bibinfo{title}{Deep residual learning for image recognition}, in:
  \bibinfo{booktitle}{2016 IEEE Conference on Computer Vision and Pattern
  Recognition (CVPR)}, pp. \bibinfo{pages}{770--778}.
\newblock \DOIprefix\doi{10.1109/CVPR.2016.90}.
\bibitem[{Horn and Schunck(1981)}]{optical_flow}
\bibinfo{author}{Horn, B.}, \bibinfo{author}{Schunck, B.},
  \bibinfo{year}{1981}.
\newblock \bibinfo{title}{Determining optical flow}.
\newblock \bibinfo{journal}{Artificial Intelligence} \bibinfo{volume}{17},
  \bibinfo{pages}{185--203}.
\newblock \DOIprefix\doi{10.1016/0004-3702(81)90024-2}.
\bibitem[{Kalman(1960)}]{kalman_filter}
\bibinfo{author}{Kalman, R.E.}, \bibinfo{year}{1960}.
\newblock \bibinfo{title}{A new approach to linear filtering and prediction
  problems}.
\newblock \bibinfo{journal}{Transactions of the ASME--Journal of Basic
  Engineering} \bibinfo{volume}{82}, \bibinfo{pages}{35--45}.
\bibitem[{Kim et~al.(2018)Kim, Kook, Sun, Kang and Ko}]{fpn}
\bibinfo{author}{Kim, S.W.}, \bibinfo{author}{Kook, H.K.},
  \bibinfo{author}{Sun, J.Y.}, \bibinfo{author}{Kang, M.C.},
  \bibinfo{author}{Ko, S.J.}, \bibinfo{year}{2018}.
\newblock \bibinfo{title}{Parallel feature pyramid network for object
  detection}, in: \bibinfo{editor}{Ferrari, V.}, \bibinfo{editor}{Hebert, M.},
  \bibinfo{editor}{Sminchisescu, C.}, \bibinfo{editor}{Weiss, Y.} (Eds.),
  \bibinfo{booktitle}{Computer Vision -- ECCV 2018},
  \bibinfo{publisher}{Springer International Publishing},
  \bibinfo{address}{Cham}. pp. \bibinfo{pages}{239--256}.
\bibitem[{Kuhn(1955)}]{algoritmo_ungherese}
\bibinfo{author}{Kuhn, H.W.}, \bibinfo{year}{1955}.
\newblock \bibinfo{title}{The hungarian method for the assignment problem}.
\newblock \bibinfo{journal}{Naval Research Logistics Quarterly}
  \bibinfo{volume}{2}, \bibinfo{pages}{83--97}.
\newblock \URLprefix
  \url{https://onlinelibrary.wiley.com/doi/abs/10.1002/nav.3800020109},
  \DOIprefix\doi{https://doi.org/10.1002/nav.3800020109},
  \href{http://arxiv.org/abs/https://onlinelibrary.wiley.com/doi/pdf/10.1002/nav.3800020109}{\tt
  arXiv:https://onlinelibrary.wiley.com/doi/pdf/10.1002/nav.3800020109}.
\bibitem[{Leal{-}Taix{\'{e}}(2014)}]{tracking_by_detection2}
\bibinfo{author}{Leal{-}Taix{\'{e}}, L.}, \bibinfo{year}{2014}.
\newblock \bibinfo{title}{Multiple object tracking with context awareness}.
\newblock \bibinfo{journal}{CoRR} \bibinfo{volume}{abs/1411.7935}.
\newblock \URLprefix \url{http://arxiv.org/abs/1411.7935},
  \href{http://arxiv.org/abs/1411.7935}{\tt arXiv:1411.7935}.
\bibitem[{Leal-Taix{\'e} et~al.(2015)Leal-Taix{\'e}, Milan, Reid, Roth and
  Schindler}]{mot15}
\bibinfo{author}{Leal-Taix{\'e}, L.}, \bibinfo{author}{Milan, A.},
  \bibinfo{author}{Reid, I.D.}, \bibinfo{author}{Roth, S.},
  \bibinfo{author}{Schindler, K.}, \bibinfo{year}{2015}.
\newblock \bibinfo{title}{Motchallenge 2015: Towards a benchmark for
  multi-target tracking}.
\newblock \bibinfo{journal}{ArXiv} \bibinfo{volume}{abs/1504.01942}.
\bibitem[{Lei et~al.(2015)Lei, Di and Jun-long}]{aereospace}
\bibinfo{author}{Lei, Q.}, \bibinfo{author}{Di, Z.}, \bibinfo{author}{Jun-long,
  L.}, \bibinfo{year}{2015}.
\newblock \bibinfo{title}{Tracking for near space nonballistic target based on
  several filter algorithms}, in: \bibinfo{booktitle}{2015 34th Chinese Control
  Conference (CCC)}, pp. \bibinfo{pages}{4997--5002}.
\newblock \DOIprefix\doi{10.1109/ChiCC.2015.7260417}.
\bibitem[{Lin et~al.(2017)Lin, Goyal, Girshick, He and Doll{\'a}r}]{focal_loss}
\bibinfo{author}{Lin, T.Y.}, \bibinfo{author}{Goyal, P.},
  \bibinfo{author}{Girshick, R.B.}, \bibinfo{author}{He, K.},
  \bibinfo{author}{Doll{\'a}r, P.}, \bibinfo{year}{2017}.
\newblock \bibinfo{title}{Focal loss for dense object detection}.
\newblock \bibinfo{journal}{2017 IEEE International Conference on Computer
  Vision (ICCV)} , \bibinfo{pages}{2999--3007}.
\bibitem[{Liu et~al.(2022)Liu, Mao, Wu, Feichtenhofer, Darrell and
  Xie}]{convnext}
\bibinfo{author}{Liu, Z.}, \bibinfo{author}{Mao, H.}, \bibinfo{author}{Wu, C.},
  \bibinfo{author}{Feichtenhofer, C.}, \bibinfo{author}{Darrell, T.},
  \bibinfo{author}{Xie, S.}, \bibinfo{year}{2022}.
\newblock \bibinfo{title}{A convnet for the 2020s}.
\newblock \bibinfo{journal}{CoRR} \bibinfo{volume}{abs/2201.03545}.
\newblock \URLprefix \url{https://arxiv.org/abs/2201.03545},
  \href{http://arxiv.org/abs/2201.03545}{\tt arXiv:2201.03545}.
\bibitem[{Luo et~al.(2017)Luo, Xing, Milan, Zhang, Liu, Zhao and
  Kim}]{mot_review}
\bibinfo{author}{Luo, W.}, \bibinfo{author}{Xing, J.}, \bibinfo{author}{Milan,
  A.}, \bibinfo{author}{Zhang, X.}, \bibinfo{author}{Liu, W.},
  \bibinfo{author}{Zhao, X.}, \bibinfo{author}{Kim, T.K.},
  \bibinfo{year}{2017}.
\newblock \bibinfo{title}{Multiple object tracking: A literature review}.
\newblock \bibinfo{journal}{Artificial Intelligence} \bibinfo{volume}{293}.
\newblock \DOIprefix\doi{10.1016/j.artint.2020.103448}.
\bibitem[{Maratea and Ferone(2019)}]{Ferone}
\bibinfo{author}{Maratea, A.}, \bibinfo{author}{Ferone, A.},
  \bibinfo{year}{2019}.
\newblock \bibinfo{title}{Deep neural networks and explainable machine
  learning}, in: \bibinfo{editor}{Full{\'e}r, R.}, \bibinfo{editor}{Giove, S.},
  \bibinfo{editor}{Masulli, F.} (Eds.), \bibinfo{booktitle}{Fuzzy Logic and
  Applications}, \bibinfo{publisher}{Springer International Publishing},
  \bibinfo{address}{Cham}. pp. \bibinfo{pages}{253--256}.
\bibitem[{Miao et~al.(2016)Miao, Zou, Li, Zhang, Wang and
  He}]{video_surveillance}
\bibinfo{author}{Miao, Z.}, \bibinfo{author}{Zou, S.}, \bibinfo{author}{Li,
  Y.}, \bibinfo{author}{Zhang, X.}, \bibinfo{author}{Wang, J.},
  \bibinfo{author}{He, M.}, \bibinfo{year}{2016}.
\newblock \bibinfo{title}{Intelligent video surveillance system based on moving
  object detection and tracking}.
\newblock \bibinfo{journal}{DEStech Transactions on Engineering and Technology
  Research} \DOIprefix\doi{10.12783/dtetr/iect2016/3765}.
\bibitem[{O'Shea and Nash(2015)}]{cnn_intro}
\bibinfo{author}{O'Shea, K.}, \bibinfo{author}{Nash, R.}, \bibinfo{year}{2015}.
\newblock \bibinfo{title}{An introduction to convolutional neural networks}.
\newblock \bibinfo{journal}{ArXiv e-prints} .
\bibitem[{Redmon et~al.(2015)Redmon, Divvala, Girshick and Farhadi}]{yolo}
\bibinfo{author}{Redmon, J.}, \bibinfo{author}{Divvala, S.K.},
  \bibinfo{author}{Girshick, R.B.}, \bibinfo{author}{Farhadi, A.},
  \bibinfo{year}{2015}.
\newblock \bibinfo{title}{You only look once: Unified, real-time object
  detection}.
\newblock \bibinfo{journal}{CoRR} \bibinfo{volume}{abs/1506.02640}.
\newblock \URLprefix \url{http://arxiv.org/abs/1506.02640},
  \href{http://arxiv.org/abs/1506.02640}{\tt arXiv:1506.02640}.
\bibitem[{Ren et~al.(2015)Ren, He, Girshick and Sun}]{faster_rcnn}
\bibinfo{author}{Ren, S.}, \bibinfo{author}{He, K.}, \bibinfo{author}{Girshick,
  R.}, \bibinfo{author}{Sun, J.}, \bibinfo{year}{2015}.
\newblock \bibinfo{title}{Faster r-cnn: Towards real-time object detection with
  region proposal networks}.
\newblock \bibinfo{journal}{IEEE Transactions on Pattern Analysis and Machine
  Intelligence} \bibinfo{volume}{39}.
\newblock \DOIprefix\doi{10.1109/TPAMI.2016.2577031}.
\bibitem[{Rosten and Drummond(2006)}]{fast}
\bibinfo{author}{Rosten, E.}, \bibinfo{author}{Drummond, T.},
  \bibinfo{year}{2006}.
\newblock \bibinfo{title}{Machine learning for high-speed corner detection},
  in: \bibinfo{editor}{Leonardis, A.}, \bibinfo{editor}{Bischof, H.},
  \bibinfo{editor}{Pinz, A.} (Eds.), \bibinfo{booktitle}{Computer Vision --
  ECCV 2006}, \bibinfo{publisher}{Springer Berlin Heidelberg},
  \bibinfo{address}{Berlin, Heidelberg}. pp. \bibinfo{pages}{430--443}.
\bibitem[{Shao et~al.(2018)Shao, Zhao, Li, Xiao, Yu, Zhang and
  Sun}]{crowd_human_dataset}
\bibinfo{author}{Shao, S.}, \bibinfo{author}{Zhao, Z.}, \bibinfo{author}{Li,
  B.}, \bibinfo{author}{Xiao, T.}, \bibinfo{author}{Yu, G.},
  \bibinfo{author}{Zhang, X.}, \bibinfo{author}{Sun, J.}, \bibinfo{year}{2018}.
\newblock \bibinfo{title}{Crowdhuman: {A} benchmark for detecting human in a
  crowd}.
\newblock \bibinfo{journal}{CoRR} \bibinfo{volume}{abs/1805.00123}.
\newblock \URLprefix \url{http://arxiv.org/abs/1805.00123},
  \href{http://arxiv.org/abs/1805.00123}{\tt arXiv:1805.00123}.
\bibitem[{Shuai et~al.(2020)Shuai, Berneshawi, Modolo and Tighe}]{trackrcnn}
\bibinfo{author}{Shuai, B.}, \bibinfo{author}{Berneshawi, A.G.},
  \bibinfo{author}{Modolo, D.}, \bibinfo{author}{Tighe, J.},
  \bibinfo{year}{2020}.
\newblock \bibinfo{title}{Multi-object tracking with siamese track-rcnn}.
\newblock \bibinfo{journal}{CoRR} \bibinfo{volume}{abs/2004.07786}.
\newblock \URLprefix \url{https://arxiv.org/abs/2004.07786},
  \href{http://arxiv.org/abs/2004.07786}{\tt arXiv:2004.07786}.
\bibitem[{Singh et~al.(2019)Singh, Gupta and
  Singh}]{human_computer_interaction}
\bibinfo{author}{Singh, S.}, \bibinfo{author}{Gupta, A.},
  \bibinfo{author}{Singh, T.}, \bibinfo{year}{2019}.
\newblock \bibinfo{title}{Computer vision based hand gesture recognition a
  survey}.
\newblock \bibinfo{journal}{International Journal of Computer Sciences and
  Engineering} \bibinfo{volume}{7}, \bibinfo{pages}{507--515}.
\newblock \DOIprefix\doi{10.26438/ijcse/v7i5.507515}.
\bibitem[{Sun et~al.(2021)Sun, Cao, Jiang, Yuan, Bai, Kitani and
  Luo}]{sun2022dance}
\bibinfo{author}{Sun, P.}, \bibinfo{author}{Cao, J.}, \bibinfo{author}{Jiang,
  Y.}, \bibinfo{author}{Yuan, Z.}, \bibinfo{author}{Bai, S.},
  \bibinfo{author}{Kitani, K.}, \bibinfo{author}{Luo, P.},
  \bibinfo{year}{2021}.
\newblock \bibinfo{title}{Dancetrack: Multi-object tracking in uniform
  appearance and diverse motion}.
\newblock \bibinfo{journal}{CoRR} \bibinfo{volume}{abs/2111.14690}.
\newblock \URLprefix \url{https://arxiv.org/abs/2111.14690},
  \href{http://arxiv.org/abs/2111.14690}{\tt arXiv:2111.14690}.
\bibitem[{Tai et~al.(2004)Tai, Tseng, Lin and Song}]{intelligence_monitoring}
\bibinfo{author}{Tai, J.C.}, \bibinfo{author}{Tseng, S.T.},
  \bibinfo{author}{Lin, C.P.}, \bibinfo{author}{Song, K.T.},
  \bibinfo{year}{2004}.
\newblock \bibinfo{title}{Real-time image tracking for automatic traffic
  monitoring and enforcement applications}.
\newblock \bibinfo{journal}{Image and Vision Computing} \bibinfo{volume}{22},
  \bibinfo{pages}{485--501}.
\newblock \URLprefix
  \url{https://www.sciencedirect.com/science/article/pii/S0262885603002439},
  \DOIprefix\doi{10.1016/j.imavis.2003.12.001}.
\bibitem[{Tan and Le(2019)}]{efficientnet}
\bibinfo{author}{Tan, M.}, \bibinfo{author}{Le, Q.V.}, \bibinfo{year}{2019}.
\newblock \bibinfo{title}{Efficientnet: Rethinking model scaling for
  convolutional neural networks}.
\newblock \bibinfo{journal}{ArXiv} \bibinfo{volume}{abs/1905.11946}.
\bibitem[{Tang et~al.(2019)Tang, Naphade, Liu, Yang, Birchfield, Wang, Kumar,
  Anastasiu and Hwang}]{automatic_driving}
\bibinfo{author}{Tang, Z.}, \bibinfo{author}{Naphade, M.},
  \bibinfo{author}{Liu, M.}, \bibinfo{author}{Yang, X.},
  \bibinfo{author}{Birchfield, S.}, \bibinfo{author}{Wang, S.},
  \bibinfo{author}{Kumar, R.}, \bibinfo{author}{Anastasiu, D.C.},
  \bibinfo{author}{Hwang, J.}, \bibinfo{year}{2019}.
\newblock \bibinfo{title}{Cityflow: {A} city-scale benchmark for multi-target
  multi-camera vehicle tracking and re-identification}.
\newblock \bibinfo{journal}{CoRR} \bibinfo{volume}{abs/1903.09254}.
\newblock \URLprefix \url{http://arxiv.org/abs/1903.09254},
  \href{http://arxiv.org/abs/1903.09254}{\tt arXiv:1903.09254}.
\bibitem[{Vaswani et~al.(2017)Vaswani, Shazeer, Parmar, Uszkoreit, Jones,
  Gomez, Kaiser and Polosukhin}]{transformers}
\bibinfo{author}{Vaswani, A.}, \bibinfo{author}{Shazeer, N.},
  \bibinfo{author}{Parmar, N.}, \bibinfo{author}{Uszkoreit, J.},
  \bibinfo{author}{Jones, L.}, \bibinfo{author}{Gomez, A.N.},
  \bibinfo{author}{Kaiser, L.}, \bibinfo{author}{Polosukhin, I.},
  \bibinfo{year}{2017}.
\newblock \bibinfo{title}{Attention is all you need}.
\newblock \bibinfo{journal}{CoRR} \bibinfo{volume}{abs/1706.03762}.
\newblock \URLprefix \url{http://arxiv.org/abs/1706.03762},
  \href{http://arxiv.org/abs/1706.03762}{\tt arXiv:1706.03762}.
\bibitem[{Wojke et~al.(2017)Wojke, Bewley and Paulus}]{deep_sort}
\bibinfo{author}{Wojke, N.}, \bibinfo{author}{Bewley, A.},
  \bibinfo{author}{Paulus, D.}, \bibinfo{year}{2017}.
\newblock \bibinfo{title}{Simple online and realtime tracking with a deep
  association metric}, in: \bibinfo{booktitle}{2017 IEEE International
  Conference on Image Processing (ICIP)}, pp. \bibinfo{pages}{3645--3649}.
\newblock \DOIprefix\doi{10.1109/ICIP.2017.8296962}.
\bibitem[{Wu et~al.(2013)Wu, Lim and Yang}]{challenge_tracking}
\bibinfo{author}{Wu, Y.}, \bibinfo{author}{Lim, J.}, \bibinfo{author}{Yang,
  M.H.}, \bibinfo{year}{2013}.
\newblock \bibinfo{title}{Online object tracking: A benchmark}, in:
  \bibinfo{booktitle}{Proceedings of the IEEE Conference on Computer Vision and
  Pattern Recognition (CVPR)}, pp. \bibinfo{pages}{2411--2418}.
\bibitem[{Xiao et~al.(2016)Xiao, Li, Wang, Lin and Wang}]{cuhs}
\bibinfo{author}{Xiao, T.}, \bibinfo{author}{Li, S.}, \bibinfo{author}{Wang,
  B.}, \bibinfo{author}{Lin, L.}, \bibinfo{author}{Wang, X.},
  \bibinfo{year}{2016}.
\newblock \bibinfo{title}{End-to-end deep learning for person search}.
\newblock \bibinfo{journal}{CoRR} \bibinfo{volume}{abs/1604.01850}.
\newblock \URLprefix \url{http://arxiv.org/abs/1604.01850},
  \href{http://arxiv.org/abs/1604.01850}{\tt arXiv:1604.01850}.
\bibitem[{Zhang et~al.(2016)Zhang, Liu, Ma and Fu}]{siamese_network}
\bibinfo{author}{Zhang, C.}, \bibinfo{author}{Liu, W.}, \bibinfo{author}{Ma,
  H.}, \bibinfo{author}{Fu, H.}, \bibinfo{year}{2016}.
\newblock \bibinfo{title}{Siamese neural network based gait recognition for
  human identification}, in: \bibinfo{booktitle}{2016 IEEE International
  Conference on Acoustics, Speech and Signal Processing (ICASSP)}, pp.
  \bibinfo{pages}{2832--2836}.
\newblock \DOIprefix\doi{10.1109/ICASSP.2016.7472194}.
\bibitem[{Zhang et~al.(2017)Zhang, Benenson and Schiele}]{city_persons_dataset}
\bibinfo{author}{Zhang, S.}, \bibinfo{author}{Benenson, R.},
  \bibinfo{author}{Schiele, B.}, \bibinfo{year}{2017}.
\newblock \bibinfo{title}{Citypersons: A diverse dataset for pedestrian
  detection}.
\newblock \bibinfo{journal}{2017 IEEE Conference on Computer Vision and Pattern
  Recognition (CVPR)} , \bibinfo{pages}{4457--4465}.
\bibitem[{Zhang et~al.(2021a)Zhang, Sun, Jiang, Yu, Yuan, Luo, Liu and
  Wang}]{byte_track}
\bibinfo{author}{Zhang, Y.}, \bibinfo{author}{Sun, P.}, \bibinfo{author}{Jiang,
  Y.}, \bibinfo{author}{Yu, D.}, \bibinfo{author}{Yuan, Z.},
  \bibinfo{author}{Luo, P.}, \bibinfo{author}{Liu, W.}, \bibinfo{author}{Wang,
  X.}, \bibinfo{year}{2021}a.
\newblock \bibinfo{title}{Bytetrack: Multi-object tracking by associating every
  detection box}.
\newblock \bibinfo{journal}{CoRR} \bibinfo{volume}{abs/2110.06864}.
\newblock \URLprefix \url{https://arxiv.org/abs/2110.06864},
  \href{http://arxiv.org/abs/2110.06864}{\tt arXiv:2110.06864}.
\bibitem[{Zhang et~al.(2021b)Zhang, Wang, Wang, Zeng and Liu}]{fairmot_paper}
\bibinfo{author}{Zhang, Y.}, \bibinfo{author}{Wang, C.}, \bibinfo{author}{Wang,
  X.}, \bibinfo{author}{Zeng, W.}, \bibinfo{author}{Liu, W.},
  \bibinfo{year}{2021}b.
\newblock \bibinfo{title}{Fairmot: On the fairness of detection and
  re-identification in multiple object tracking}.
\newblock \bibinfo{journal}{International Journal of Computer Vision}
  \bibinfo{volume}{129}, \bibinfo{pages}{1--19}.
\newblock \DOIprefix\doi{10.1007/s11263-021-01513-4}.
\bibitem[{Zheng et~al.(2016)Zheng, Zhang, Sun, Chandraker and
  Tian}]{prw_dataset}
\bibinfo{author}{Zheng, L.}, \bibinfo{author}{Zhang, H.}, \bibinfo{author}{Sun,
  S.}, \bibinfo{author}{Chandraker, M.}, \bibinfo{author}{Tian, Q.},
  \bibinfo{year}{2016}.
\newblock \bibinfo{title}{Person re-identification in the wild}.
\newblock \bibinfo{journal}{CoRR} \bibinfo{volume}{abs/1604.02531}.
\newblock \URLprefix \url{http://arxiv.org/abs/1604.02531},
  \href{http://arxiv.org/abs/1604.02531}{\tt arXiv:1604.02531}.

\end{thebibliography}


\end{document}